\documentclass[]{article}

\usepackage{booktabs}
\usepackage[table]{xcolor}
\usepackage{graphicx}
\usepackage{mathtools}
\usepackage{stmaryrd}
\usepackage{multicol}
\usepackage{multirow}
\usepackage{subcaption}
\usepackage{enumitem}
\usepackage{cite}
\usepackage{pifont}

\newcommand{\cmark}{\ding{51}}
\newcommand{\xmark}{\ding{55}}

\title{PARDINUS: Weakly supervised discarding of photo-trapping empty images based on autoencoders}

\author{David de la Rosa \and
Antonio J Rivera \and
María J del Jesus \and
Francisco Charte\thanks{Computer Science Department,  Universidad de Ja\'en, Campus Las Lagunillas, 23071 Ja\'en, Spain.}
}

\begin{document}
\maketitle              

\begin{abstract}

Photo-trapping cameras are widely employed for wildlife monitoring. Those cameras take photographs when motion is detected to capture images where animals appear. A significant portion of these images are empty - no wildlife appears in the image. Filtering out those images is not a trivial task since it requires hours of manual work from biologists. Therefore, there is a notable interest in automating this task. Automatic discarding of empty photo-trapping images is still an open field in the area of Machine Learning. Existing solutions often rely on state-of-the-art supervised convolutional neural networks that require the annotation of the images in the training phase. PARDINUS (Weakly suPervised discARDINg of photo-trapping empty images based on aUtoencoderS) is constructed on the foundation of weakly supervised learning and proves that this approach equals or even surpasses other fully supervised methods that require further labeling work.

\end{abstract}

\section{Introduction} \label{Introduction}

The conservation and sustainable management of natural environments is a vital task. Forests not only serve as carbon sinks but they are also home to a diversity of plants and animal species.

Effective forest management requires a broad knowledge of the species that inhabit these areas, as their behaviors, habits and interactions with the environment. To achieve this, modern technology has provided us with useful tools known as camera traps.

Camera traps \cite{fototrampeo} are unobtrusive devices equipped with motion sensors that allow us to capture images of wildlife without human interference, providing much information about the natural environment and animal species.

Nevertheless, the analysis of the images taken by camera traps is a time-consuming task. It requires professionals able to recognize animal species and relate the knowledge gained from different images to control the status of the species. On top of that, a new problem appears, the empty images. Empty images are those that do not depict any animals, showing the landscape of the images. An animal that passed too quickly in front of the camera, the sunlight or even the heat may trigger the motion sensor and capture an empty image. Images of this kind are of no interest to scientists since they are useless for species monitoring. 

In the past, it was common to manually filter empty images by inspecting the photographs and discarding those that have no animal presence. Generally, the number of empty images exceeds the number of images that contain wildlife. This fact increases significantly the cost of this task, as more empty images imply more time and human resources to dedicate to this labor. In the last years, another approach has emerged \cite{aiFototrampeo}: automatic discarding, this is, the use of software capable of determining by itself whether or not an image is empty. Artificial Intelligence (AI) \cite{aiRef} and, more specifically, Deep Learning (DL) \cite{dlRef}, a subfield of Machine Learning (ML) that focuses on training neural networks with multiple layers to derive higher-level abstractions from vast amounts of data, discovering complex patterns and features, constitutes the predominant approach for this task. Nevertheless, most of the algorithms such as \cite{Megadetector}, \cite{animalscanner} or \cite{mlwic2}, rely on supervised learning, training state-of-the-art convolutional neural networks to detect empty images. Those networks require the images to be annotated, which is again a time-consuming task that involves reviewing and annotating each training image.

The present work aims to provide a weakly supervised solution for the photo-trapping empty image filtering problem. The major part of the algorithm is constructed on unsupervised techniques, in particular clustering and autoencoders. Only the last piece of the algorithm, a classifier, follows a supervised approach, thus reducing the amount of annotated data required to train the models. 

This algorithm also showcases its potential to outperform traditional supervised techniques, surpassing the performance of existing approaches.

The document is structured as follows: section \ref{Preliminaries} addresses the state of the art in the field of photo trapping image processing, indicating the main weaknesses we found. Section \ref{Proposal} shows the workflow of PARDINUS (Weakly suPervised discARDINg of photo-trapping empty images based on aUtoencoderS), the proposed system, whose details and analysis of the architecture and parameters are presented in section \ref{Experimentation}. The performance of PARDINUS and a comparison with other state-of-the-art solutions is covered in section \ref{Results}. Lastly, section \ref{Conclusions} shows the conclusions derived from the study upon its completion.

\section{Preliminaries}\label{Preliminaries}

As indicated in the previous section, it is possible to extract a large amount of knowledge from photo-trapping images. However, the manual processing of those images is a hard, repetitive and time-consuming task. For this reason, many solutions can be found in the literature to achieve this objective using AI and techniques based on computer vision \cite{cvRef}. The proposals, which will be explained below, can be divided into two main groups: studies on the classification and detection of empty images and studies that focus on both the detection of empty images and the classification of the species in the photographs.

% Descartar imágenes vacías

\subsection{Detection of empty images}

The first common task that often arises in this context is the identification of empty images. This task is essential for biologists as empty images provide no valuable information and should be discarded. In the literature, two algorithmic proposals based on computer vision, named Zilong \cite{zilong} and Sherlock \cite{Sherlock}, have been proposed to address this problem.

Zilong assumes that every time the camera detects motion, it captures a sequence of photographs. The software analyzes each pack of photographs and calculates the difference among the images. This is done in a two-step process. Being \textit{$p_1$} and \textit{$p_2$} two continuous shot images of the same dimensions, they first calculate the number of pixels whose difference exceeds a preset threshold $\alpha$. If the number of different pixels is larger than the user-input criteria $c_1$, it means that there is motion and that the images are not empty.

Sherlock takes a similar approach. In this case, instead of calculating the difference among the images, it creates a background image as the median value of each pixel. Then it calculates the difference between each image and the background image to determine the presence of animals. This is done by randomly sampling pixels to find disturbances, sets of pixels that are different from the background image.

According to studies such as \cite{embeddedSystem} and \cite{nospcimen}, DL has shown significant success in solving this problem.

The study \cite{embeddedSystem} proposes to insert a DL module for detecting empty images in the camera itself. It avoids the subsequent discarding phase as only non-empty images are saved. Researchers tested two types of models trained on Caltech Camera Traps \cite{terraIncognita} and Snapshot Serengeti \cite{snapshot} datasets adapted for binary classification. On the one hand, classifier models such as MobileNetV2 \cite{mobilenetv2} and EfficientNet \cite{efficientnet} are able to analyze the content of the image and classify it as containing animals or not. On the other hand, detector models such as SSDLite \cite{mobilenetv2} with a MobileNetV2 as backbone and EfficientNet, that not only determine if an image contains animals but also where the animals are located in the non-empty frames. They proved that detection models outperform classifiers but their superior inference latency can limit their use on edge devices.

NOSpcimen algorithm \cite{nospcimen} was based on the idea that a set of AEs specifically trained to reconstruct empty images, each AE specialized in images with similar features, would not be able to accurately reconstruct images where animals appear. On this basis, after the AEs training, the reconstruction error between the original images and the posterior reconstruction made by the AEs is calculated. With these errors and knowing which one corresponds to an empty frame and which one to an animal image, it is possible to train a neural network specialized in the recognition of non-empty images taking as input the reconstruction error. Results showed that, although NOSpcimen was surpassed by other state-of-the-art methods, it was a valid approach for the discarding of empty images, achieving an area under the ROC curve of 0.955 and an F1-score of 0.891 for the dataset that was used in the study. Moreover, due to its weakly supervised nature, the required annotation work is considerably less than in other proposals.

% Descartar vacías y clasificar

\subsection{Detection of empty images and species classification}
Species classification is another common approach in photo-trapping camera processing \cite{villa2017towards}. Nevertheless, many proposals combine these two tasks of species classification and empty image identification, since both fit well as classification problems. Two variants can be found in this case.

\textbf{The first methodology} is to address this task as a single multiclass classification problem. Instead of training a model to classify images into one of the $N$ classes, where $N$ is the number of species that we want to detect, one extra class is added for images where there is no animal presence. The main proposals that follow this methodology are described below.

One of the best-known proposals in this area is Megadetector \cite{Megadetector}, a tool for classifying images as containing animals, vehicles, people or if it is empty. The latest version uses the YOLOv5 \cite{yolov5} network, which allows image classification with high processing speed. This model was trained on bounding boxes from a variety of ecosystems with both private and public data. Because of that, Megadetector not only classifies images but also includes bounding boxes to indicate where animals (or other classes of objects) are located in the images.

Animal Scanner \cite{animalscanner} assumes that photographs are captured in a sequence every time the camera detects motion. For each sequence, it analyzes the images individually. First, the image is divided into smaller blocks that are classified as foreground or background to distinguish the moving objects in the foreground. To determine which blocks contain moving animals, they compare each block with the other co-located blocks of the sequence. These foreground regions need to be classified into human, animal or empty by an AI model. They tested three different classifiers: Bag of Visual Word \cite{bagofvisualwords}, AlexNet \cite{alexnet} and AlexNet-96, a variation that takes as input images of 96x96 pixels. As Megadetector, Animal Scanner also provides a user-friendly desktop application.

Convolutional neural networks considered state-of-the-art have been also tested to solve this problem by treating it as a multiclass classification in which one of the classes determines that the image does not show any animal. In the research presented in \cite{mlvshuman}, a comparison between manual classification and the classification performed by ResNet50 network \cite{resnet50} is presented. The findings reveal that fully automated species labels can be comparable to expert labels, particularly in estimating the number of species, discerning activity patterns and determining occupancy (the estimation that a site is occupied by a species given it was not detected). 

Six state-of-the-art networks, DenseNet201 \cite{densenet}, Inception-ResNet-V3 \cite{inceptionv3}, InceptionV3 \cite{inceptionv3}, NASNetMobile \cite{nasnet}, MobileNetV2 \cite{mobilenetv2} and Xception \cite{xception}, were also tested and compared for camera trap images multiclass classification in \cite{schneider2020three}. They trained the networks with 47\,279 images labeled across 55 classes, compared the performance and detected that DenseNet201 performed best on both the training set, with 95.6\% top-1 accuracy, and the test set, with 68.7\% top-1 accuracy.

Some studies as \cite{terraIncognita} attempted to verify the generalization of state-of-the-art models, ResNet-101 \cite{resnet18} and InceptionResNet-v2 with atrous convolution \cite{inceptionastrous} to novel environments on which they have not been trained. Those models were trained to classify the animal species that appear in the images, including a specific class for empty images. Their experiments showed that state-of-the-art algorithms show decent performance when tested on the same location where they were trained, but poor generalization to new locations.

The insertion of the DL module into the camera is also possible in this case as indicated in the article \cite{iot}. Researchers use Raspberry Pi and models based on the ResNet-18 architecture \cite{resnet18} to detect the species of animals that appear in the images and ghost images, which refers to empty images or other anomalies.

Lastly, some researchers focus on the detection and recognition of a specific species in camera traps. It is the case of \cite{Desertbighornsheep}, which specializes in the detection of Desert bighorn sheep among other species in the same area or empty images. They tried well-known supervised machine learning techniques such as k-nearest neighbors, support vector machines, neural networks and Naive Bayes.

\textbf{The second methodology} is to create a two-phase classification scheme. The first phase focuses on the detection of empty images. A binary classification model trained specifically for this task is used to detect and discard, as appropriate, the empty images. The rest of the images, those where animals appear, are classified in the second phase using a multiclass classification model to predict which species appear in the image. As there are two different models, the tools that will be presented can be used to detect and discard empty images, classify animal species or both.

Wildlife Insights \cite{wildlife} is a web platform that can be used to upload data to Google Cloud and access AI models trained to automatically classify camera trap images. It has two main artificial intelligence modules. The first, called \textit{wildlife identification service}, is a multiclass classification convolutional neural network using pre-trained image embedding from Inception \cite{inceptionv3}. They fine-tuned the model using a labeled dataset with over 35 million images covering 1295 different species. This module can be used for both filtering out empty images or classifying animals. The second module, the \textit{analytics engine module}, provides statistics and analyses such as occupancy, density or species diversity.

MLWIC2 \cite{mlwic2} is an R package for the processing of photo-trapping cameras. It contains two pre-trained models based on ResNet18 architecture \cite{resnet18}. Both models were trained with 3 million camera trap images from 18 studies in 10 states across the United States of America. On the one hand, the \textit{species model} can identify 58 species and empty images.  On the other hand, the \textit{empty-animal model} distinguishes between empty and non-empty images, regardless of the species of the animal. The models performed well on some out-of-sample datasets, achieving an accuracy between 36\% and 91\% on the species model and an accuracy between 01\% and 94\% on the empty-animal model. It also provides functionality for users to train a new model using their images.

The study \cite{norouzzadeh2018automatically} trained and tested nine state-of-the-art convolutional neural networks on the Snapshot Serengeti dataset \cite{snapshot}: AlexNet \cite{alexnet}, NiN \cite{nin}, VGG \cite{vgg}, GoogLeNet and five different variations of ResNet \cite{resnet18}. For the task of detecting images that contain animals, VGG performed the best. For the species classification, the best results were given by ResNet. Also, the study includes additional phases for counting the number of animals in the images and attributes that indicate the animal behavior, as if it is moving, sleeping, eating, etc.

Transfer learning is also a valid approach in this area. In the article \cite{willi2019identifying}, researchers used two ResNet18 models \cite{resnet18}, one for empty image detection and another for species classification, trained on the Snapshot Serengeti dataset \cite{snapshot} and applied transfer-learning by copying and freezing the weights of the convolutional layers and only re-learned the weights of the fully connected layers. They proved that, in most cases, transfer learning surpassed the accuracy of models trained from scratch, even if species in the base model differ from the species in the target dataset.

\subsection{Summary}

As discussed in the background section, there are two groups of studies in the field of photo-trapping image processing: detection of empty images and a combination of detecting empty images and animal species classification. PARDINUS will fully focus on the detection of empty photo trap images, becoming part of the first group.

\begin{table}[ht!]
  \centering
  \small
  \begin{tabular}{l c@{\hskip 8pt} c@{\hskip 8pt} c@{\hskip 8pt} r} 
    \toprule
      \textbf{Name} & \textbf{SML} & \textbf{CV} & \textbf{UML} & \textbf{Reference} \\
    \midrule
      Zilong & \xmark & \cmark & \xmark & \cite{zilong} \\
      Sherlock & \xmark & \cmark & \xmark & \cite{Sherlock}\\
      Cunha\_2021 & \cmark & \xmark & \xmark & \cite{embeddedSystem} \\
      Megadetector & \cmark & \xmark & \xmark & \cite{Megadetector} \\
      Animal Scanner & \cmark & \cmark & \xmark & \cite{animalscanner} \\
      MBaza & \cmark & \xmark & \xmark & \cite{mlvshuman} \\
      Schneider\_2020 & \cmark & \xmark & \xmark & \cite{schneider2020three} \\
      Terra Incognita & \cmark & \xmark & \xmark & \cite{terraIncognita} \\
      Zualkernan\_2020 & \cmark & \xmark & \xmark & \cite{iot} \\
      Vargas\_2021 & \cmark & \xmark & \xmark & \cite{Desertbighornsheep} \\
      Wildlife Insights & \cmark & \xmark & \xmark & \cite{wildlife} \\
      MLWIC2 & \cmark & \xmark & \xmark & \cite{mlwic2} \\
      Norouzzadeh\_2017 & \cmark & \xmark & \xmark & \cite{norouzzadeh2018automatically} \\
      Willi\_2019 & \cmark & \xmark & \xmark & \cite{willi2019identifying} \\
      NOSpcimen & \cmark & \xmark & \cmark & \cite{nospcimen} \\
    \bottomrule
    
  \end{tabular}
  \caption{Classification of studies that address the discarding empty photo trap images problem as they use supervised machine learning (SML), computer vision techniques (CV) or unsupervised machine learning (UML).}
  \label{Tabla.EstadoArte}
\end{table}

Table \ref{Tabla.EstadoArte} represents the state-of-the-art methods in the field of photo-trapping analysis, providing insights into the techniques and approaches employed to address this problem and classifying them as they use supervised ML, unsupervised ML or computer vision techniques. 

As stated, most of the studies that fully or partially address this problem use supervised ML techniques to train state-of-the-art convolutional neural networks able to process and recognize empty images. Just three of the considered studies, Zilong, Sherlock and Animal Scanner, changed the methodology and applied computer vision techniques for this purpose. None of them except NOSpcimen, our previous work, introduces weakly supervised ML techniques in the software and works independently of the temporal order in the images.

On the other hand, some proposals require the images to be arranged in a temporal order to compare the photographs taken in the same sequence. Although this increases the knowledge that the model takes to make the prediction, it may pose a problem when using datasets whose images are not temporally arranged, as is our case.

For those reasons, to reduce the needed time to annotate the images for the training phase, this study will focus on weakly supervised DL approaches for the discarding of empty images. Furthermore, the algorithm will be designed to operate independently of the temporal arrangement of images.

\section{Our proposal: PARDINUS system} \label{Proposal}

PARDINUS is the proposed algorithm to solve the empty image filtering problem. 

\begin{figure}[h!]
    \centering
    \includegraphics[width=0.6\textwidth]{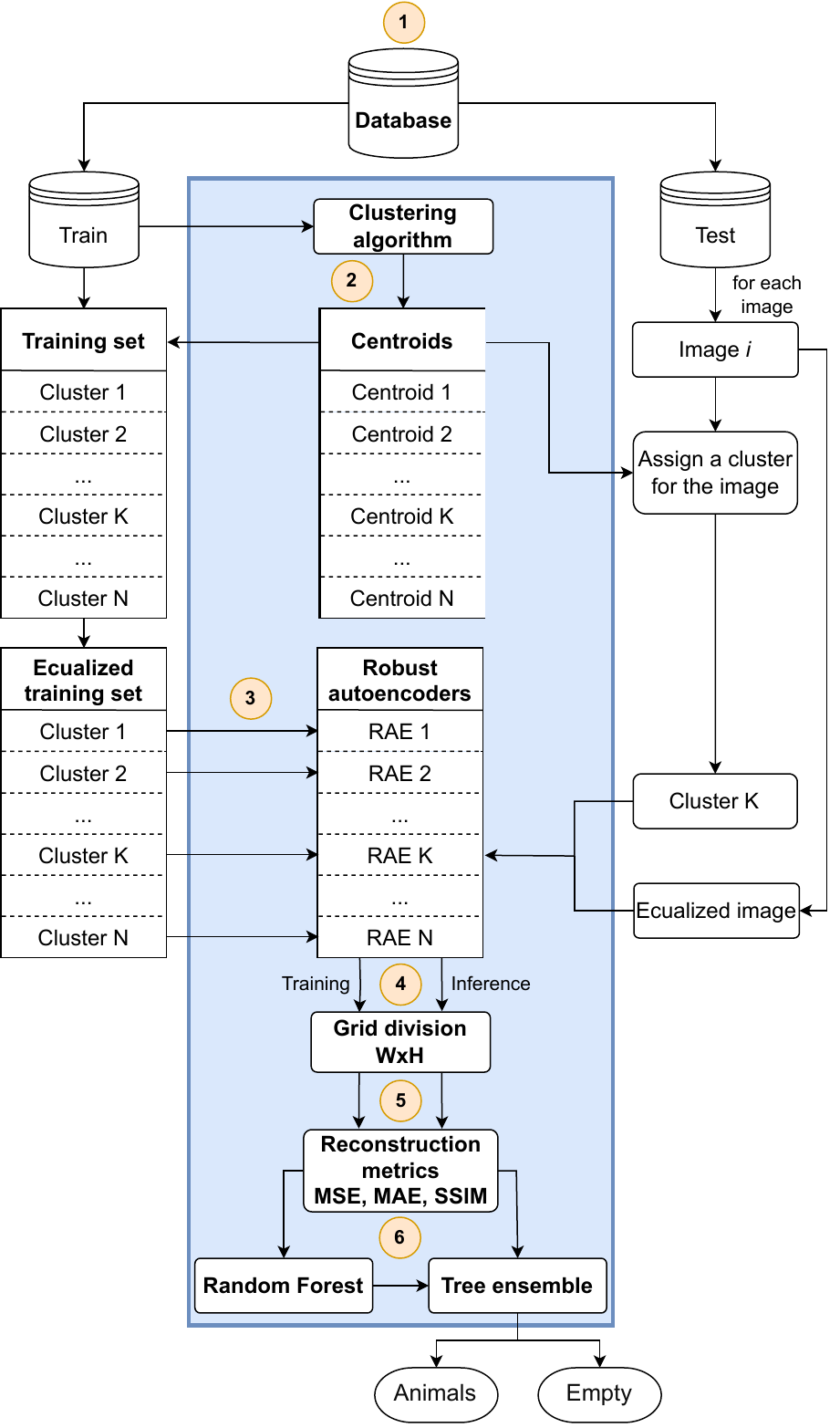}
    \caption{Diagram that represents the structure of PARDINUS. It illustrates the training and inference process for empty photo-trapping images detection.}
    \label{Fig.esquemaGeneral}
\end{figure}

As stated in section \ref{Preliminaries}, the objective of this paper is to define a weakly supervised workflow for the discarding of empty images. On this basis, AEs \cite{autoencoderCharte}, unsupervised DL techniques, are the mainstay of this solution. PARDINUS has four main phases: RGB image clustering to form $N$ groups of similar images, equalized image reconstruction made by the AE specialized in the cluster associated with the image, by first dividing the image into $W \times H$ blocks and calculating the difference between each block of the original image and its associated block in the reconstructed image, and image classification based on the reconstruction error. Its diagram is presented in figure \ref{Fig.esquemaGeneral}. 

The training process follows the next steps:

\begin{enumerate}
    \item Train - test split is done.
    \item A clustering algorithm is trained on RGB images to derive the centroids of each group.
    \item The clustered and equalized training set is employed to train $N$ robust autoencoders, one for each cluster of images.
    \item A division of $W \times H$ blocks is made for each original and reconstructed image.
    \item The reconstruction error metrics between the original and reconstructed images are calculated.
    \item Those metrics are used to train an RF model, resulting in a tree ensemble optimized for classification.
\end{enumerate}

As it is logical, the inference process follows the same steps. For each image, we assign a cluster using the centroids. The equalized image is reconstructed by the associated RAE and divided into $W \times H$ blocks, for which reconstruction metrics are calculated. Passing those metrics and the cluster identifier as input to the tree ensemble, the system makes the prediction.

This algorithm was derived through extensive experimentation detailed in section \ref{Experimentation}. Further information is provided in the following subsections.

\subsection{Unsupervised training of the clustering algorithm}
\begin{figure}[h!]
    \centering
    \begin{subfigure}{.22\textwidth}
        \centering
        \includegraphics[width=\textwidth]{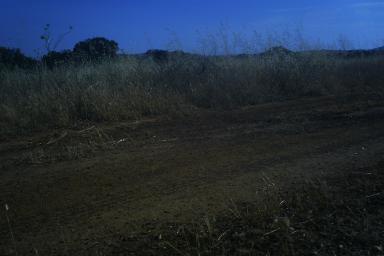}
    \end{subfigure}
    \hspace{1mm}
    \begin{subfigure}{.22\textwidth}
        \centering
        \includegraphics[width=\textwidth]{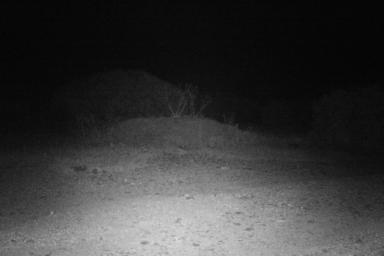}
    \end{subfigure}
    \hspace{1mm}
    \begin{subfigure}{.22\textwidth}
        \centering
        \includegraphics[width=\textwidth]{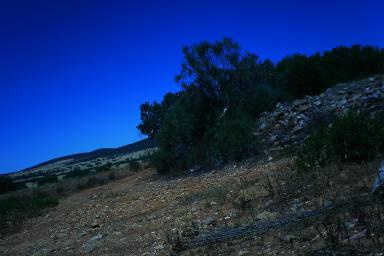}
    \end{subfigure}
    \hspace{1mm}
    \begin{subfigure}{.22\textwidth}
        \centering
        \includegraphics[width=\textwidth]{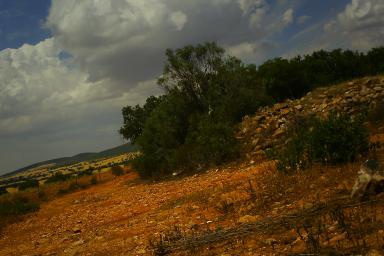}
    \end{subfigure}

    \vspace{2mm}

    \begin{subfigure}{.22\textwidth}
        \centering
        \includegraphics[width=\textwidth]{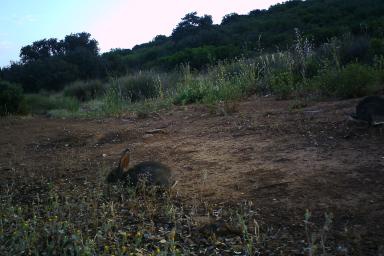}
    \end{subfigure}
    \hspace{1mm}
    \begin{subfigure}{.22\textwidth}
        \centering
        \includegraphics[width=\textwidth]{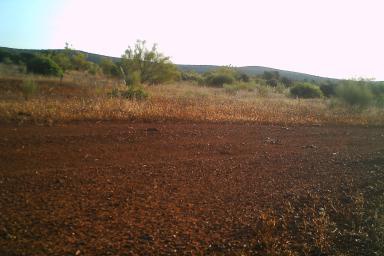}
    \end{subfigure}
    \hspace{1mm}
    \begin{subfigure}{.22\textwidth}
        \centering
        \includegraphics[width=\textwidth]{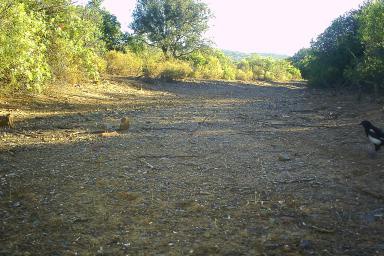}
    \end{subfigure}
    \hspace{1mm}
    \begin{subfigure}{.22\textwidth}
        \centering
        \includegraphics[width=\textwidth]{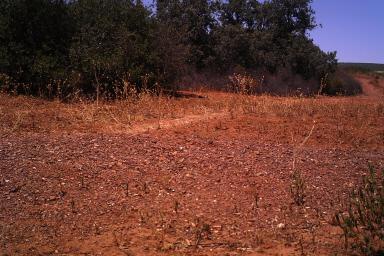}
    \end{subfigure}
    
    \caption{Representation of some of the images taken from the WWF21 dataset.}
    \label{Fig.analisisVisualWWF21}
\end{figure}

The photo-trapping cameras that captured the images forming our dataset, details of which will be presented in section \ref{Experimentation}, are located in natural environments. As depicted in the figure \ref{Fig.analisisVisualWWF21}, those areas are very dynamic. Depending on the camera location, we can find images taken in forest areas with heavy or low vegetation, dry areas, images captured in larger or smaller areas, etc. Even images taken in the same location but at different times of the year differ from each other.

In addition, the time of day also affects the appearance of the image. While daytime images are usually highly variable, night images are taken in black-and-white shades which means that, regardless of the location, they are similar to each other.

All that knowledge has to be learned by a DL model capable of extracting the features of the images to determine if there is animal presence or not, which is a complex task.

For this reason, we propose the training dataset segmentation to create $N$ groups of images (stages $1$ and $2$ in the figure \ref{Fig.esquemaGeneral}). To do so, we apply the Lloyd K-Means algorithm \cite{kmeansLloyd} on RGB images, which assigns each image to a specific cluster $K$ based on the distance of the image to the centroids and selects the closest group. The number of clusters to form, $N$, depends on the dataset. If the dataset contains images with similar features, a low number of clusters may be correct. In contrast, if the images are disparate, it may be appropriate to increase the number of clusters to form.

\subsection{Unsupervised training of AEs}
The main idea is to detect anomalies, that is, images where animals appear, by using AE \cite{autoencoderCharte}. An AE is a neural network that usually follows an unsupervised workflow. Its objective is to reconstruct the inputs at its output. By modifying the architecture of the network we can get a new data representation in the encoding layer, often of smaller dimensions. Many variants of the original AE architecture can be found in the literature \cite{autoencoderCharte} such as denoising autoencoders, sparse autoencoders or variational autoencoders. In this case, robust autoencoders (RAE) \cite{RobustAE}, AEs that use correntropy-based loss function \cite{Correntropy}, will be used because of their higher tolerance to noisy inputs.

To use the RAE for the detection of empty images, we train $N$ RAEs, one for each cluster, to only reconstruct this kind of image (stage 3 in the figure \ref{Fig.esquemaGeneral}). RAEs will take as input equalized images.

Training RAEs only with empty images makes the reconstruction of images where animals appear of poorer quality. A high difference between the original and its reconstruction means that the RAE has not been able to rebuild the image properly because, as it has no knowledge about how to rebuild the part where the animal appears, the reconstruction of the animal is just a blur, leading to high error. 

This occurrence can be detected using a classifier trained to identify images where an animal appears by analyzing the dissimilarities between the original image and its reconstruction generated by the RAE.

\subsection{Calculation of the reconstruction errors}
As mentioned, the next step is to compare the original and the reconstructed images. To measure the difference, we use three metrics: mean squared error (MSE), mean absolute error (MAE) and structural similarity (SSIM). Its equations are \ref{Eq.MSE}, \ref{Eq.MAE} and \ref{Eq.SSIM}.

\begin{equation}
    \label{Eq.MSE}
    MSE = \frac{1}{m * n}\sum_{i=0}^{m-1}\sum_{j=0}^{n-1}[Y(i,j) - \hat{Y}(i,j)]^2
\end{equation}

\begin{equation}
    \label{Eq.MAE}
    MAE = \frac{1}{m * n}\sum_{i=0}^{m-1}\sum_{j=0}^{n-1}|Y(i,j) - \hat{Y}(i,j)|
\end{equation}

\begin{equation}
    \label{Eq.SSIM}
    SSIM(x,y) = \frac{(2\mu_x\mu_y + c_1)(2\sigma_{xy} + c_2)}{(\mu_x^2 + \mu_y^2 + c_1)(\sigma_x^2 + \sigma_y^2 + c_2)}
\end{equation}

Those equations use the following nomenclature. For MSE and MAE, $m$ is the image width, $n$ the image height, $Y$ the original image and $\hat{Y}$ the reconstructed image, and for SSIM, $x$ is the original image, $y$ the reconstructed image, $\mu$ is the pixel sample mean of the image, $\sigma^2$ is the variance of the image, $\sigma$ the covariance of the images, $c_1 = (k_1L)^2$ and $c_2 = (k_2L)^2$, where $L$ is the dynamic range of the pixel values, $k_1 = 0.01$ and $k_2 = 0.03$ by default.

Nevertheless, when dealing with photographs containing animals, we often encounter an additional challenge in the images where the animal occupies only a small portion of it. In such cases, the conventional evaluation metrics described before may not accurately capture the fidelity of the reconstruction process since the RAE model excels at reconstructing the background of the image while struggling with the animal itself. Consequently, when the animal is only a small portion of the image, the RAE model performs poorly only in that specific area, resulting in a very low reconstruction error. This can potentially lead the classifier to classify it as an empty image.

To address this challenge, we decided to partition both the original and reconstructed images into $W \times H$ blocks, organized in a grid format of $W$ blocks in width and $H$ blocks in height (stage 4 of the figure \ref{Fig.esquemaGeneral}). This strategy serves a critical purpose. It allows us to obtain localized error measurements calculating MSE, MAE and SSIM for each block (stage 5 of the figure \ref{Fig.esquemaGeneral}). This means that we can easily identify specific regions within the image where the reconstruction is poor, thus highlighting possible areas where an animal may be present even if it is small. The number of blocks depends on the images of the dataset. If the animal occupies a large part of the image, the number of blocks to be created will be smaller. On the other hand, if the animal occupies a small portion of the image, as in our case, the number of blocks should be larger.

Moreover, we observed variations in the reconstruction quality among different clusters, regardless of the presence or absence of animals. The images captured at night are a good example. That kind of image is taken in gray tones: the background is completely dark and we only see the area that the flash illuminates. As the background, which is half of the image, is just black, it makes the reconstruction very accurate, independently of the presence or absence of animals. Conversely, the images taken in an environment with dense vegetation are harder to reconstruct since shrubs and trees have subtle details that are not easy to detect for the RAE. This is a trouble for the classifier. A night image where an animal appears may have a high error in the context of its cluster, but this error may be low in the context of another cluster. Therefore, we add the cluster identifier with the $W \times H$ error metrics to enable the classifier to know the difference between clusters.

\subsection{Classification of the reconstruction errors}
The last step is to predict if an image is empty or not by taking as input the $W \times H$ reconstruction errors, three per block, together with the cluster identifier.

To achieve this, we trained a Random Forest machine-learning algorithm \cite{randomForest} (RF) to build a tree ensemble capable of distinguishing between errors from empty images and errors from non-empty images (stage 6 of the figure \ref{Fig.esquemaGeneral}). This allows us to determine the presence or absence of animals in the photo-trapping images.

\section{Analysis of the architecture and parameters of the algorithm} \label{Experimentation}

This section covers the experimentation that led to the model presented in the previous section. It starts with subsections \ref{dt} and \ref{qm}, which provide some information about the dataset and indicate which quality metrics have been used to measure the model performance.

Subsection \ref{cl} covers the experimentation related to clustering. We aim to analyze the optimal method for clustering to facilitate the reconstruction of empty images by AEs. To achieve this goal, this subsection provides detailed experiments using various clustering approaches, including clustering above image histograms, RGB images or gray images, among other techniques. In addition, the number of clusters that the algorithm creates is also discussed.

We explain the experimentation related to AEs and the classification algorithm in section \ref{aecl}. Starting from a base algorithm, our focus lies in exploring the diverse landscape of AE types present in the literature. With numerous variations in the base architecture, our goal is to identify the type and internal structure of AE that best adapts to our dataset. Additionally, we delve into various methods of calculating the reconstruction errors between original and reconstructed images. Lastly, we undertake a comprehensive study to determine which machine learning algorithm exhibits better performance.

The last subsection \ref{pardresults} exhibits the results of the chosen algorithm, PARDINUS, presented in section \ref{Proposal}.

It is important to note that, given the substantial number of variables involved in this study, the process has been conducted incrementally. We begin with a base algorithm and architecture, detailed in subsection \ref{baseArch}, and conduct experiments in blocks to analyze specific aspects of the algorithm. Once the optimal configuration is determined, we proceed to the next block of experiments, using the architecture obtained from the previous step as the base.

\subsection{WWF21 dataset} \label{dt}
The dataset employed in this research, hereinafter referred to as WWF21, comprises a collection of 45\,728 images collected from photo-trapping cameras installed in various natural parks in Spain by WWF Spain. In the dataset, we can find two types of images. On the one hand, a majority of \textit{empty} images, specifically 37\,503 images devoid of visible wildlife, will be used to train the clustering algorithms and the AEs. On the other hand, 8\,225 \textit{animal} images that contain wildlife species and will be used to train the last block of the system, the classifier algorithm.

The dataset includes images taken at eighteen different locations from eastern Sierra Morena (Spain). All images in the dataset are in RGB (Red, Green, Blue) format, which is a standard color representation format for digital images, and have dimensions of 2\,560 pixels wide and 1\,920 pixels high.

To prepare data for model input, some preprocessing is necessary. A smaller image size is valid for a neural network and even beneficial since it considerably reduces the amount of memory required and training time. For that reason, we decreased the dimensions of the images from 2560 pixels in width and 1920 pixels in height to 384 pixels in width and 256 pixels in height.

Moreover, a train-test split is required in order to ensure proper evaluation of the model. We follow the proportion 60-20-20, 60\% images for training, 20\% for validation and 20\% for testing.

\subsection{Quality metrics} \label{qm}

To know how the model behaves, performance evaluation is paramount. Many metrics serve as quantitative measures of the model's performance.

Among the metrics, the Area Under the Curve (AUC) stands out as a robust indicator of overall model performance, capturing the trade-off between the true positive rate (sensitivity) and the true negative rate (specificity). A high AUC value implies that the model can correctly distinguish positive and negative instances.

Furthermore, in the context of this problem, it is crucial to minimize False Negatives (FN), that is, images where animals appear classified as empty, as they represent valuable images that might be erroneously discarded. By taking care of this metric, we ensure that valuable data is not lost during the classification process.

In addition to those essential metrics, we calculate two additional ones: accuracy (ACC) and False Positives (FP). On the one hand, ACC serves as a generic and well-known metric to check how many samples have been correctly classified. Nevertheless, due to the disbalance of the dataset, the accuracy metric is not as good as AUC, which does take into account the inhomogeneous distribution of the two types of images. On the other hand, FP allows us to check how many empty images have been classified as images where animals appear. A higher FP implies more human work, as empty images that have not been discarded are non-useful.

\subsection{Clustering study} \label{cl}

There are multiple usable implementations of the K-Means algorithm. In this case, the chosen one is the Scikit-learn \cite{sklearn} version. Its ease of use and the possibility of modifying training parameters as the number of clusters made this a good choice.

\begin{figure}[h!]
    \centering
    \includegraphics[width=0.7\textwidth]{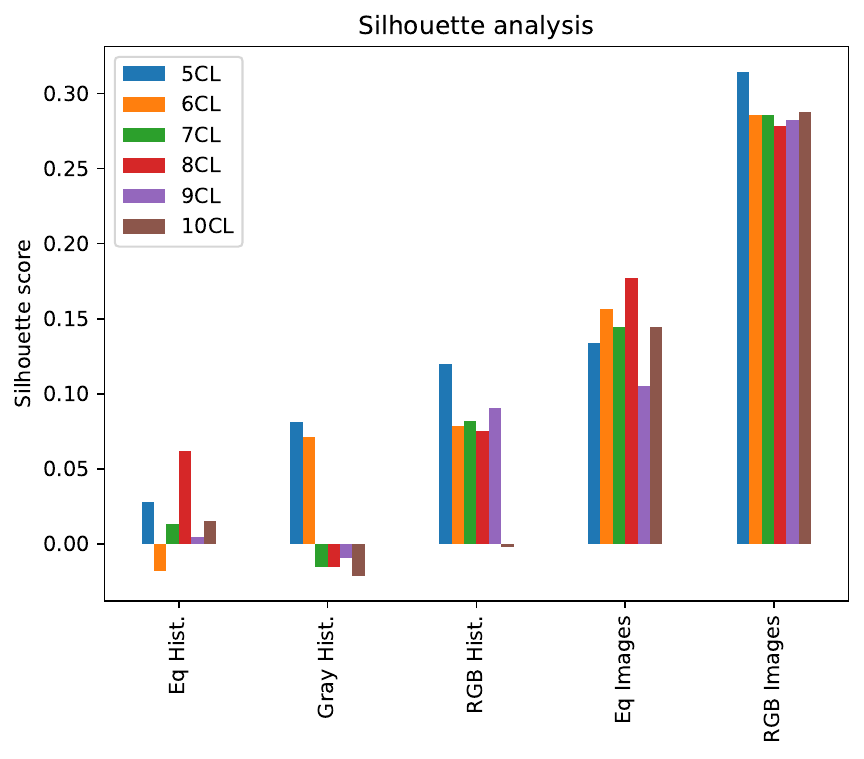}
    \caption{Silhouette score for five types of clustering, being Eq Hist. equalized histogram, Gray Hist. gray tones histogram, RGB Hist. red, green, blue histogram, Eq Images equalized images, RGB Images red, green, blue images.}
    \label{Fig.silhouette}
\end{figure}

The first step is to determine how the clustering is to be done. NOSpcimen relied on the histogram of the image since it is a good method to describe an image. In this case, we have created five experiments to verify if this is the best choice: clustering on RGB histograms (NOSpcimen), equalized histograms, black and white histograms, equalized images and RGB images. To measure the clustering quality we use the average silhouette score \cite{silhouette}, a measure of how similar an object is to its cluster compared to other clusters. As can be seen in figure \ref{Fig.silhouette}, grouping the images using the image itself instead of its histogram gives the best results. NOSpcimen's method occupies the third position, proving that, although it yields a logical clustering, using the RGB images emerges as the optimal choice. Equalized histogram and gray tones histogram remain in the last positions.

\begin{figure}[h!]
    \centering
    \includegraphics[width=0.75\textwidth]{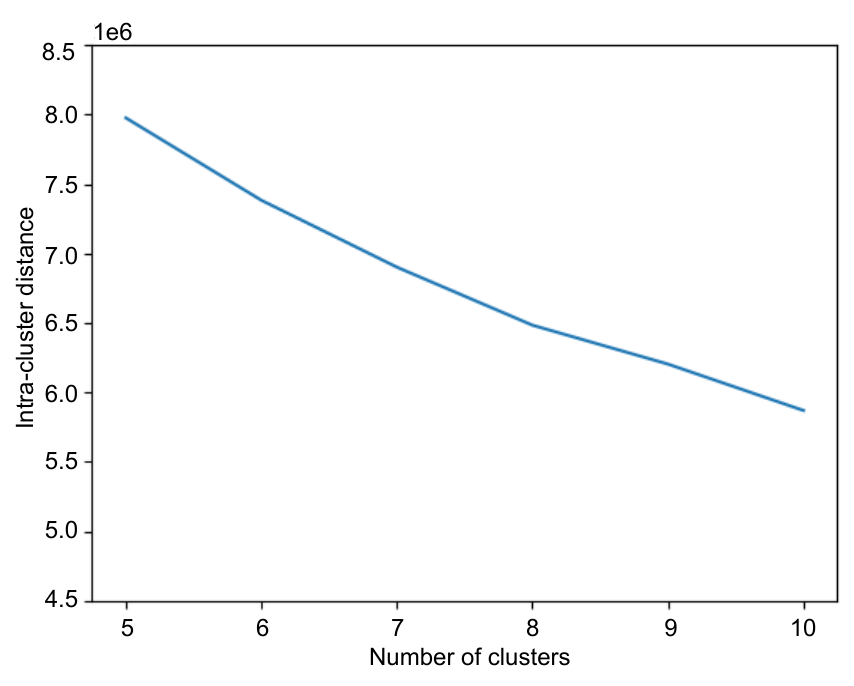}
    \caption{Intra-cluster distance over the number of clusters.}
    \label{Fig.intracluster}
\end{figure}

Next, we have to set the number of groups that the algorithm will create. A low number of clusters may not be enough for the diversity of the dataset, while a high number of clusters leads to some clusters having very few images, which can be a drawback when training AEs. In this experiment, the number of clusters has been tested with values ranging from five to ten. As can be seen in figure \ref{Fig.intracluster}, beyond seven and eight clusters, the intra-cluster distance starts to decrease more gradually. Taking into consideration that with a high number of clusters images containing animals begin to have limited representation in some groups, a fact that occurs from eight groups onward, and noting that the Silhouette coefficient is slightly higher with seven clusters (according to figure \ref{Fig.silhouette}), we have decided to form seven clusters for this dataset.

\subsection{Image reconstruction and classification study} \label{aecl}

This subsection covers the experimentation related to the AE models and the classifier, taking into account image format, dataset balancing, and AE and classifier architecture. The base architecture of the AEs and classifier from which we start the experimentation is also detailed in this subsection.

\subsubsection{Base AE and classifier architecture}  \label{baseArch} \hfill\\

Before starting with the experimentation related to AEs and the classifier, it is necessary to detail the base architecture of AEs and the classifier from which we started this study.

\begin{figure}[h!]
    \centering
    \includegraphics[width=0.8\textwidth]{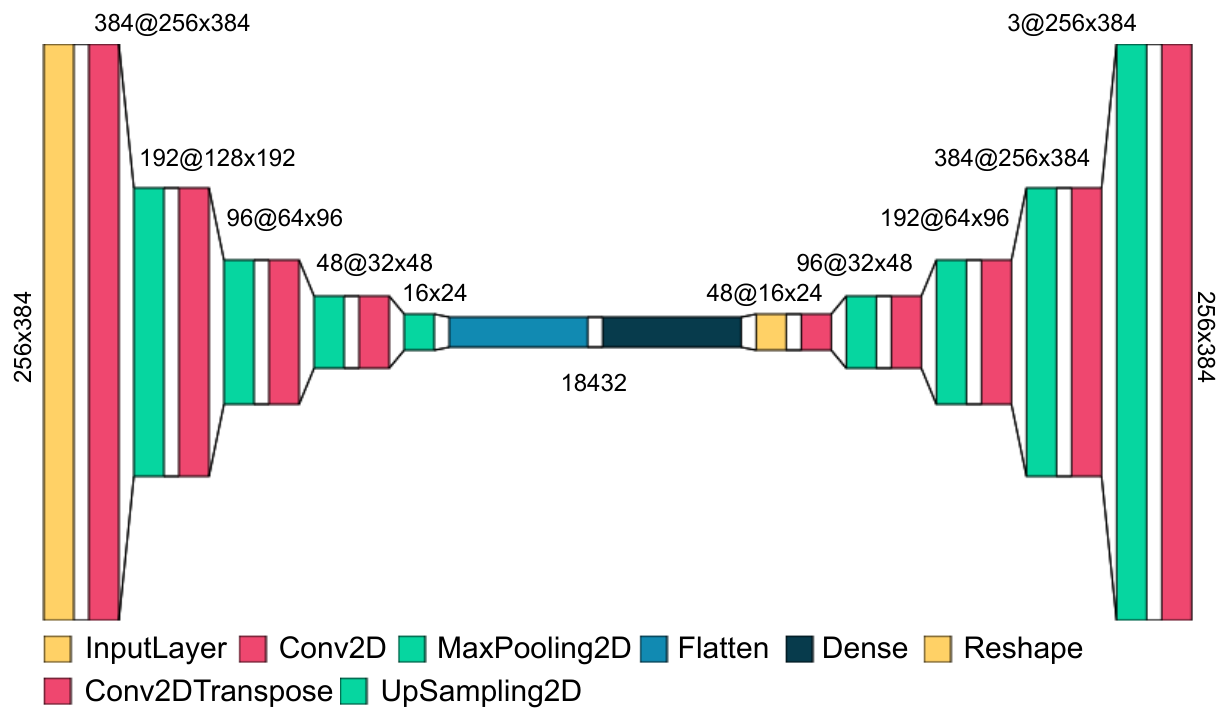}
    \caption{Architecture of the original RAE model.}
    \label{Fig.arquitectura}
\end{figure}

Figure \ref{Fig.arquitectura} represents the original architecture of the AEs. 

It is a symmetrical architecture formed by four packs of convolutional and max-pooling layers on the encoder, a dense layer of 18\,432 neurons, and four packs of transposed convolutional and up-sampling layers in the decoder. It takes as input three-channel images with dimensions 256 pixels high and 384 pixels wide. The output of the network is the reconstructed image, with the same dimensions as the input. We use RAE as the initial variety of autoencoder.

As stated in the previous paragraphs, we created seven clusters based on RGB images. Therefore, we trained seven RAEs, one for each cluster. RAEs take as input equalized images. Each of the RAEs was trained over $70$ epoch and a batch size of $16$ images.

The base algorithm employs the MLP as a reconstruction error classifier. This model has two hidden dense layers, each with $20$ neurons, and an output softmax layer consisting of two neurons that to facilitate the classification.

As indicated at the start of this section, this initial algorithm will be modified in each block of experiments.

\subsubsection{Image format prior to reconstruction} \hfill\\

The image format of the image when passed to the AE as input can be decisive. We hypothesize that equalized images can improve contrast in the images and highlight the animal presence. 

To prove it, we created two experiments. In one of them, we use RGB images and reconstruct it using the base architecture of the RAE. Next, we calculate the MSE, MAE, and SSIM for both empty and animal original and reconstructed images. The other experiment follows the same steps but uses equalized images. Then we calculate the Diff.MSE, Diff.MAE and Diff.SSIM for both experiments following the equations \ref{eq::ecualizadas}.

\begin{equation}
\begin{aligned}
    &\textit{Diff. MSE} = |\textit{MSE}_{animal} - \textit{MSE}_{empty}| \\
    &\textit{Diff. MAE} = |\textit{MAE}_{animal} - \textit{MAE}_{empty}| \\
    &\textit{Diff. SSIM} = |\textit{SSIM}_{animal} - \textit{SSIM}_{empty}|
\end{aligned}
\label{eq::ecualizadas}
\end{equation}

A greater difference between the reconstruction of empty and non-empty images implies a greater ease in classifying the instances. As can be seen in table \ref{Tabla.Ecualizar}, the difference between the reconstruction of empty images and images where animals appear is greater in equalized images, making it the optimal choice.

\begin{table}[ht!]
  \centering
  \small
  \setlength{\tabcolsep}{5pt}
  \begin{tabular}{l c c}
    \toprule
      \textbf{Metric} & \textbf{RGB} & \textbf{Equalized} \\
      & \textbf{images} & \textbf{images} \\
    \midrule
      Diff.MSE & 0.0037 & 0.0055  \\
      Diff.MAE & 0.0178 & 0.0186 \\
      Diff.SSIM & 0.1468 & 0.1754 \\
    \bottomrule
  \end{tabular}
  \caption{The disparity in reconstruction quality metrics between empty images and images containing animals evaluated on two types of images: RGB images and equalized images.}
  \label{Tabla.Ecualizar}
\end{table}

%%%%%%%%%%%%%%%%%
\hfill \\
\subsubsection{Influence of dataset imbalance} \hfill\\

As indicated in subsection \ref{dt}, the dataset presents a class imbalance that can lead to a potential bias in the classifier model. Three different experiments have been conducted to determine the best treatment for this issue. To do so, we made the classification without applying any balancing method, balancing the dataset by removing random empty instances to equal the number of images where animals appear, and balancing the instances by cluster, that is, for each cluster, balance the number of images by removing empty images. For these experiments, we used the base AE and classifier architecture. Results are presented in table \ref{Tabla.Balanceo}.

\begin{table}[ht!]
  \centering
  \small
  \setlength{\tabcolsep}{5pt}
  \begin{tabular}{l r r r r}
    \toprule
      \textbf{Method} & \textbf{AUC} & \textbf{FN} & \textbf{FP} & \textbf{ACC} \\
      
    \midrule
      Imbalanced & 0.8359 & 0.7519 & 0.3320 & 0.8368 \\
      Balanced & 0.8297 & 0.1864 & 0.3091 & 0.7083 \\
      Balanced by cluster & 0.8169 & 0.3281 & 0.2249 & 0.7569 \\
      
    \bottomrule
  \end{tabular}
  \caption{Results using the same architecture as NOSpcimen applying balancing methods.}
  \label{Tabla.Balanceo}
\end{table}

Although results show that AUC is higher when no balancing method is applied, the FN is also higher, which causes the discarding of useful images. On the other hand, balanced outperforms balanced by cluster in terms of AUC and has the lowest number of FN. For this reason, we consider that balancing the dataset by removing random empty instances to equal the number of images where animals appear is the best method.

\subsubsection{Autoencoders and classifier} \hfill\\

The table \ref{Tabla.RecopilacionResultados} summarizes the experimentation that led us to the final model of this research concerning autoencoders and the classifier. We start the experimentation using the base architecture. The classifier is trained with the balanced dataset in all experiments.

\begin{table}[ht!]
  \centering
  \small
  \setlength{\tabcolsep}{5pt}
  \begin{tabular}{l l l l l l r r r r}
    \toprule
      \textbf{AE} & \textbf{CL} & \textbf{Grid} & \textbf{DF} & \textbf{IL} & \textbf{DO} & \textbf{AUC} & \textbf{FN} & \textbf{FP} & \textbf{ACC} \\
      
    \midrule
      RAE & MLP & - & \xmark & \xmark & - & 0.8297 & 0.1864 & 0.3091 & 0.7083 \\
      RAE & RF & - & \xmark & \xmark & - & 0.9323 & 0.0687 & 0.1893 & 0.8296 \\
      
      RAE & MLP & 6x4 & \xmark & \xmark & - & 0.9221 & 0.0741 & 0.1993 & 0.8203 \\
      RAE & MLP & 6x8 & \xmark & \xmark & - & 0.9281 & 0.2017 & \textbf{0.1096} & 0.8717 \\
      
      RAE & RF & 6x4 & \xmark & \xmark & - & 0.9677 & \textbf{0.0298} & 0.1416 & 0.8765 \\
      RAE & RF & 6x8 & \xmark & \xmark & - & 0.9672 & 0.0376 & 0.1429 & 0.8766 \\

      RAE & RF & 6x4 & \xmark & \xmark & Dense-50\% & 0.9666 & 0.0468 & 0.1340 & 0.8797 \\
      RAE & RF & 6x4 & \xmark & \xmark & Dense-20\% & 0.9680 & 0.0352 & 0.1372 & 0.8792 \\
      RAE & RF & 6x4 & \xmark & \xmark & Convs-20\% & 0.9666 & 0.0407 & 0.1356 & 0.8795 \\
      RAE & RF & 6x4 & \xmark & \cmark & - & 0.9673 & 0.0370 & 0.1372 & 0.8788 \\
      RAE & RF & 6x4 & \cmark & \xmark & - & \textbf{0.9702} & 0.0414 & 0.1348 & 0.8801 \\

      VAE & RF & 6x4 & \xmark & \xmark & - & 0.9461 & 0.0510 & 0.1763 & 0.8437 \\
      VAE & RF & 6x4 & \xmark & \cmark & - & 0.9546 & 0.0474 & 0.1601 & 0.8578 \\
      VAE & RF & 6x4 & \cmark & \xmark & - & 0.9495 & 0.0510 & 0.1600 & 0.8573 \\

      $\beta$4VAE & RF & 6x4 & \cmark & \xmark & - & 0.9361 & 0.0577 & 0.1876 & 0.8331 \\
      $\beta$15VAE & RF & 6x4 & \cmark & \xmark & - & 0.9188 & 0.0693 & 0.1993 & 0.8212 \\
      $\beta$30VAE & RF & 6x4 & \cmark & \xmark & - & 0.9153 & 0.0741 & 0.2041 & 0.8163 \\
      $\beta$50VAE & RF & 6x4 & \cmark & \xmark & - & 0.9123 & 0.0729 & 0.2055 & 0.8154 \\
      $\beta$150VAE & RF & 6x4 & \cmark & \xmark & - & 0.9130 & 0.0766 & 0.2001 & 0.8192 \\

      RAE & GB & 6x4 & \cmark & \xmark & - & 0.9350 & 0.0662 & 0.1735 & 0.8433 \\
      RAE & XGB & 6x4 & \cmark & \xmark & - & 0.9645 & 0.0395 & 0.1240 & \textbf{0.8894} \\

    \bottomrule
  \end{tabular}
  \caption{\small Experiments conducted during the research using the WWF21 dataset. The columns, from left to right, are as follows: AE, the type of autoencoder, CL, the type of classifier, Grid, the size of the grid, DF, a decrease in the number of convolutional filters, IL, an increase in the number of convolutional layers, DO, the use of dropout layers, AUC area under the ROC curve, FN false negatives, FP false positives, and ACC accuracy.}
  \label{Tabla.RecopilacionResultados}
\end{table}

The first area we conducted experimentation on was the classification model. NOSpcimen worked with MLP, but many other ML models can be used for predictions \cite{predictiveModels}. In the first two experiments, we tested the performance of RF compared to MLP. Results prove that RF significantly outperforms the results provided by MLP, achieving a one-tenth higher AUC and notably reducing both false negatives and false positives.

On the other hand, as stated in the section \ref{Proposal}, a grid division is proposed before calculating the reconstruction metrics since the animal may be only a small part of the image, making it difficult to correctly capture a high reconstruction error in non-empty images. We tested two possible divisions: 6 blocks wide and 4 or 8 blocks high. We test this grid for both MLP and RF as classifiers.

The results confirm our hypothesis. Applying a grid division of 6x4 blocks, AUC increases from $0.9323$ to $0.9677$ in the case of RF and from $0.8297$ to $0.9221$ in the case of MLP, along with an increase in ACC and a decrease in incorrect predictions. The results have not improved using the larger grid, indicating that we do not need more level of detail for this dataset.

The base architecture of the RAEs has also been tested in five experiments, adding dropout layers, increasing the number of convolutional layers by adding an extra convolutional and max-pooling layer in the encoder and their symmetrical parts in the decoder (IL), and halving the number of filters that are applied on each convolution to get a smaller central layer (DF). Nevertheless, results remain almost unchanged except when applying DF, with a slight improvement. In addition, the model size decreased from 1.27 GB to 325 MB, so we selected it as the best option.

In our experimentation, we explored different types of AE, including variational autoencoders (VAE) \cite{vae} and $\beta$-VAE \cite{betavae}. VAEs are well-known for their ability to model data distributions and generate samples from learned latent spaces, while $\beta$-VAEs introduce a regularization term that controls the disentanglement of the latent representation. However, results show that VAEs and $\beta$-VAEs do not improve the results given by RAEs in any experiment, not even when modifying the $\beta$ parameter. It indicates that, although a better disentangled latent representation is probably achieved, it does not facilitate the posterior classification.

Lastly, after confirming that altering the architecture of the AEs yielded no significant improvement and that modifying the classifier substantially improved the results, we opted to delve further into this aspect by trying two new ML models: Gradient Boosting (GB) and XGBoost (XGB). Of the two, XGB provided the best results compared with RF, improving in terms of ACC, FP and FN. Nevertheless, AUC is slightly higher when using RF. A higher AUC provides the flexibility to adjust the confidence threshold of the model according to expert requirements, ensuring that it continues operating effectively while adapting to specific needs.

\subsection{Results presentation} \label{pardresults}

In this subsection, we will derive various statistical metrics and generate graphical representations to facilitate a more comprehensive understanding of the results achieved by PARDINUS.

Along this section, an extensive study has been conducted. This study has led us to make the following decisions for PARDINUS:

\begin{itemize}
    \item The K-Means algorithm is trained with RGB images to create seven clusters.
    \item AEs take equalized images as input.
    \item The use of RAEs instead of other types of AEs
    \item The modification of the architecture of the RAEs to reduce the number of convolutional filters
    \item The dataset is balanced previous to the training of the classifier.
    \item The use of RF instead of other types of classifiers.
    \item The use of a 6x4 grid division to calculate the reconstruction errors.
\end{itemize}

\begin{figure}[h!]
    \centering
    \includegraphics[width=0.8\textwidth]{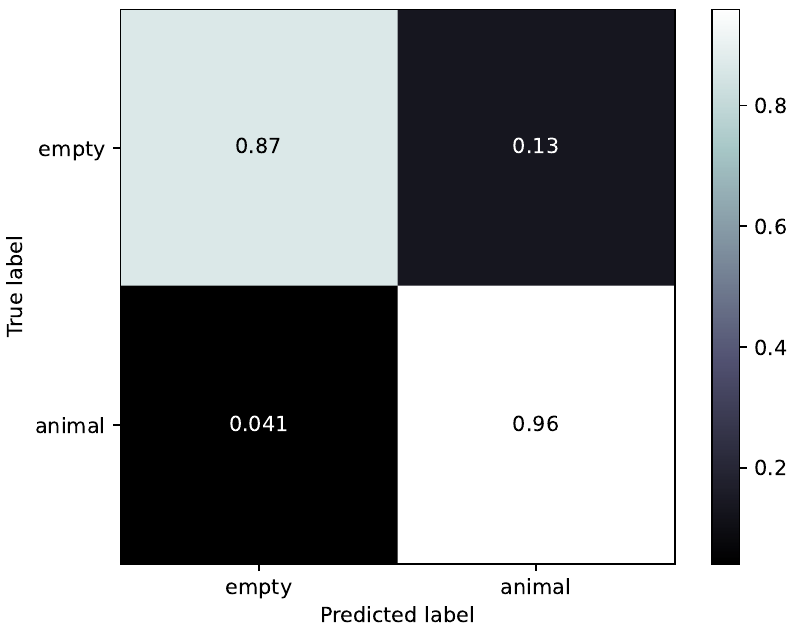}
    \caption{Confusion matrix representing the false positives, false negatives, true positives and true negatives of PARDINUS.}
    \label{Fig.confusionmatrix}
\end{figure}

Figure \ref{Fig.confusionmatrix} represents the normalized confusion matrix of PARDINUS. As can be seen, most classifications are correct since 96\% of the images where animals appear are correctly classified and only 13\% of the empty image classifications are incorrect. Those values give PARDINUS an average precision of $0.80$, an average recall of $0.91$ and an F1-score of $0.83$.

\begin{figure}[h!]
    \centering
    \includegraphics[width=0.85\textwidth]{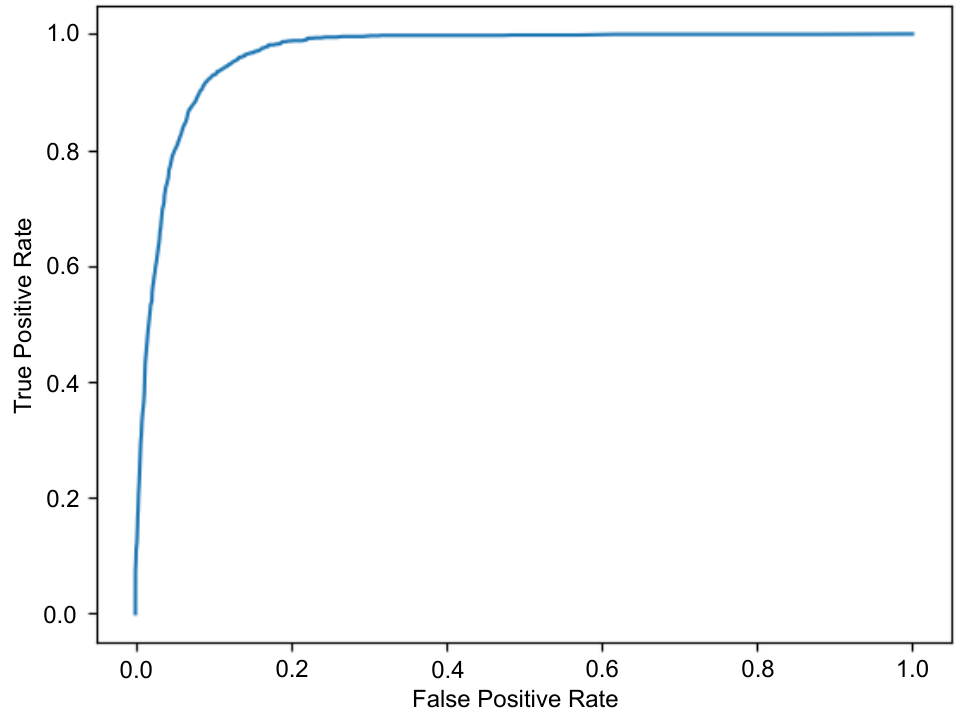}
    \caption{ROC curve of PARDINUS.}
    \label{Fig.roccurve}
\end{figure}

On the other hand, ROC curve plots the true positive rate (TPR) against the false positive rate (FPR) at different threshold settings. The perfect ROC curve forms a corner in the point [0,1], which means a false positive rate of 0 and a true positive rate of 1. The ROC curve of PARDINUS is presented in \ref{Fig.roccurve}. The quality of the ROC curve can be measured by calculating the AUC, the area under the curve. While the perfect ROC curve has an AUC of 1, PARDINUS gets an AUC of 0.9702, proving the high quality of the animal presence detection.

\section{Experimental comparison} \label{Results}

In this section, we show the outcomes of our study and conduct an analysis of PARDINUS in comparison to other existing proposals: Megadetector \cite{Megadetector}, NOSpcimen \cite{nospcimen}, MLWIC2 \cite{mlwic2}, Norouzzadeh\_2017 \cite{norouzzadeh2018automatically} and MBaza \cite{mlvshuman}.

All the experiments were performed under identical conditions, using $7500$ empty images and $1645$ non-empty images with dimensions 384 pixels wide and 256 pixels high. It is necessary to mention that all the state-of-the-art methods that have been tested had pre-trained models prepared for immediate use. Table \ref{Tabla.Comparaciones} resumes this subsection.

\begin{table}[ht!]
  \centering
  \small
  \setlength{\tabcolsep}{5pt}
  \begin{tabular}{l r r r r}
    \toprule
      \textbf{Method} & \textbf{AUC} & \textbf{FN} & \textbf{FP} & \textbf{ACC} \\
      
    \midrule
      PARDINUS & \textbf{0.9702} & \textbf{0.0414} & 0.1348 & 0.8801 \\
      Megadetector & 0.9328 & 0.2066 & \textbf{0.0341} & \textbf{0.9337} \\
      NOSpcimen & 0.8297 & 0.1864 & 0.3091 & 0.7083 \\
      MLWIC2 & 0.7448 & 0.1397 & 0.5612 & 0.5065 \\
      MBaza & 0.6614 & 0.4364 & 0.4040 & 0.5835 \\
      Norouzzadeh\_2017 & 0.6280 & 0.3212 & 0.6435 & 0.4050 \\
      
    \bottomrule
  \end{tabular}
  \caption{Results of PARDINUS and other state-of-the-art algorithms. Algorithms are sorted by AUC score.}
  \label{Tabla.Comparaciones}
\end{table}

\begin{figure}[h!]
    \centering
    \includegraphics[width=0.8\textwidth]{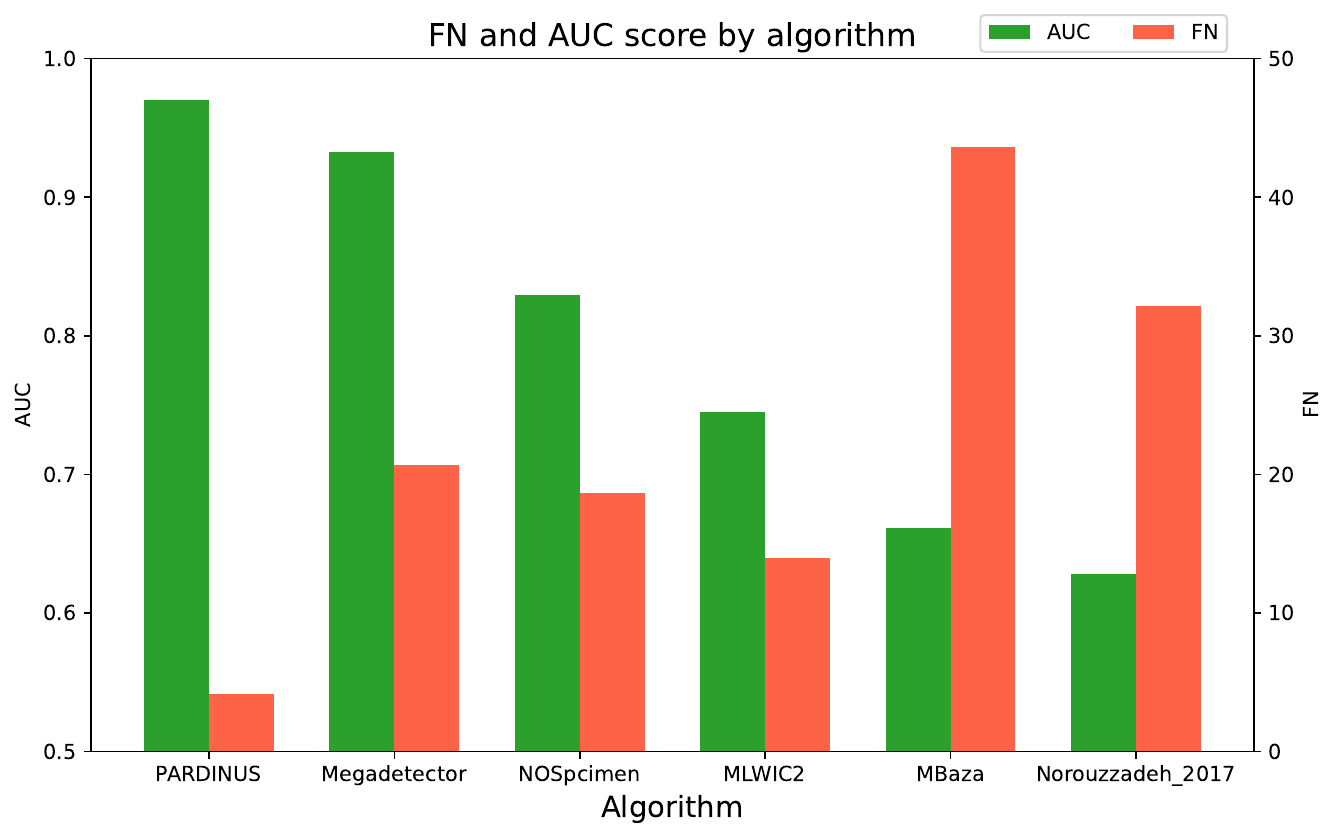}
    \caption{Bar graphic representing the FN and AUC score achieved by PARDINUS and other state-of-the-art algorithms. FN is expressed in percentages.}
    \label{Fig.comparacion}
\end{figure}

The first thing that stands out is the good results that PARDINUS achieves for the WWF21 dataset concerning other state-of-the-art models, achieving the best FN and AUC scores of all the tested methods. Figure \ref{Fig.comparacion} represents the high AUC score and low FN achieved by PARDINUS compared to the other algorithms involved in the comparison.

Even though PARDINUS is surpassed by Megadetector in terms of FP, as stated in subsection \ref{qm}, in the context of this problem FN represents valuable images that can be discarded, but FP are non-valuable images that, although they had to be removed, do not produce a loss of knowledge of the environment. For this reason, is far more important to reduce the number of FN rather than FP, which, according to the results, is precisely what PARDINUS achieves.

Regarding NOSpcimen, it is clear that PARDINUS has significantly improved its performance. In the table \ref{Tabla.Comparaciones}, we observe a notable improvement in terms of the number of FN, dropping from 20.47\% in the case of NOSpcimen to just 4.53\% in the case of PARDINUS, a trend that is also clearly reflected in figure \ref{Fig.comparacion}. This fact, along with a significant improvement in AUC, from 0.8297 to 0.9702, in ACC, from 0.7083 to 0.8801, and in FP, from 30.91\% of misclassification to 13.48\%, proves that the proposed system achieves much more accurate classifications than those obtained in the previous work. All of these improvements also prove the effectiveness of employing the grid method for calculating the reconstruction error and using the RF model instead of MLP as the classifier, alongside other crucial modifications specified in section \ref{Conclusions}.

Megadetector has also demonstrated to be a robust solution on this dataset, providing the highest accuracy, the lowest percentage of FP, and the second best AUC score, $0.9337$, $3.41$ and $0.9328$, respectively. However, it drops to the fourth position for FN score, with $20.66$\% of misclassifications. The great adaptability of Megadetector to WWF21 can be attributed to extensive training on millions of images sourced from many different locations. Despite this, Megadetector is surpassed by PARDINUS in terms of FN and AUC, the two key metrics.

In contrast, Norouzzadeh\_2017 and MBaza, designed and trained to detect animals within a specific dataset, encounter challenges when applied to our dataset, providing the worst results in terms of AUC and FN. Similarly, MLWIC2 achieved a low FN, being the second-best score, and a competitive AUC, near to NOSpcimen. However, it struggled in the context of FP since more than half of the empty images were classified as images where animals appear.

In summary, although the adaptability of Megadetector to new environments is remarkable, PARDINUS surpasses it in key metrics, showing that weakly supervised solutions like ours are also competitive in the field of empty image discarding.

\hfill \\

\section{Conclusions}\label{Conclusions}

Throughout this paper, PARDINUS, a weakly supervised deep learning algorithm for photo-traping empty images filtering has been proposed. PARDINUS has four main phases to determine if an image is empty or not. First is the clustering phase, where the RGB image is assigned to the cluster that better represents the features of the image based on the similarity between the image and the centroid of each group. Next, in the second phase, the reconstruction phase, the RAE specialized in reconstructing images within the assigned cluster reconstructs the equalized input image. In the third phase, PARDINUS calculates the reconstruction error by partitioning the original and reconstructed image into 24 blocks and obtaining MSE, MAE and SSIM for each block. The last phase includes a tree ensemble that differentiates empty and non-empty images based on the reconstruction errors and the cluster identifier.

Although NOSpcimen is our previous study and follows a similar workflow, the experimentation that led us to PARDINUS allowed us to include several improvements over NOSpcimen, such as:

\begin{itemize}
    \item The use of RGB images for the clustering phase instead of the histogram of the images. Even though the histogram of an image can describe at a general level what the image features are, we found that using the image itself for clustering provides a much more robust clustering.
    \item The use of equalized images to train the AEs instead of standard RGB images since this preprocessing emphasizes the difference between the reconstruction of animal and empty images, facilitating the posterior task of detecting the anomalies.
    \item A grid division for the original and reconstructed image to measure the reconstruction error metrics instead of calculating it for the whole image. This allows the system to properly capture small blurs in the image that could indicate the presence of small animals.
    \item The use of RF instead of MLP for the classification since, for this dataset, RF provides significantly higher results for the task of determining animal presence taking reconstruction errors as input.
\end{itemize}

Moreover, the experimentation carried out during the current study is far more extensive, covering different fields of study as the clustering, the type and number of AEs, the classifier algorithm, the architecture of the neural network or how the reconstruction metrics are calculated.

Importantly, in the present work, we conducted a thorough review of the state of the art, exploring many algorithms that address the problem through diverse methodologies, and compared our algorithm, PARDINUS, with multiple existing methods, a practice that was not employed in our previous work.

As it was analyzed, results achieved by PARDINUS overpass other approaches such as Megadetector and NOSpcimen for the WWF21 dataset. Our study underscores the potential of weakly supervised methods in achieving results comparable to larger supervised models. The weakly supervised algorithm employed in this study provides significant advantages over fully-supervised methods. Since the clustering algorithm and RAEs are constructed on unsupervised learning, they do not need labeled data for the training process. Supervised learning is only needed to train the reconstruction errors classifier. Since that model is significantly lighter than RAE models, it requires less training data hence the amount of labeled data needed to train the algorithm is small.

The practical applications of PARDINUS in wildlife conservation are evident, offering a solution for the identification and discard of empty and non-useful images, allowing scientists to considerably reduce the amount of manual work and focus on the study of images that contain wildlife.

Lastly, as future work, we are currently planning to develop a fully unsupervised method for empty image detection by eliminating the RF model and replacing it with other statistical methods. A fully unsupervised algorithm would completely remove the need for labeled data, significantly reducing the required effort during the training process since obtaining labeled data can be time-consuming and expensive.

\section*{Acknowledgements}
The research carried out in this study is part of the project ``ToSmartEADs: Towards intelligent, explainable and precise extraction of knowledge in complex problems of Data Science" financed by the Ministry of Science, Innovation and Universities with code PID2019-107793GB-I00 / AEI / 10.13039 / 501100011033.

\bibliographystyle{unsrt}

\end{document}